\documentclass[runningheads]{llncs}
\pdfoutput=1 
\usepackage{url}
\usepackage{hyperref}

\usepackage{graphicx}
\usepackage{mathtools}
\usepackage{amsmath}
\usepackage{amsfonts}
\usepackage{siunitx}
\usepackage{array}
\usepackage{multirow}
\usepackage{makecell}
\usepackage{tabularx}
\usepackage{amsmath}

\mathchardef\mhyphen="2D

\begin{document}

\title{Bayesian Inference \\ by Symbolic Model Checking\thanks{This work is funded by the ERC AdG Projekt FRAPPANT (Grant Nr. 787914).}}
\author{Bahare Salmani, \and
Joost-Pieter Katoen}
\authorrunning{B. Salmani and J.-P. Katoen}
\institute{RWTH Aachen University, Aachen, Germany \\
\email{{\{salmani,katoen\}}@cs.rwth-aachen.de}\\
}
\maketitle              % typeset the header of the contribution
\begin{abstract}
This paper applies probabilistic model checking techniques for discrete
Markov chains to inference in Bayesian networks. We present a simple
translation from Bayesian networks into tree-like Markov chains such
that inference can be reduced to computing reachability probabilities.
Using a prototypical implementation on top of the Storm model checker,
we show that symbolic data structures such as multi-terminal BDDs
(MTBDDs) are very effective to perform inference on large Bayesian
network benchmarks. We compare our result to inference using probabilistic sentential decision diagrams and vtrees, a scalable symbolic technique in AI
inference tools.
\end{abstract}
%
%
%

%%%%%%%%%%%%%%%%%%%%%%%%%%%%%%%%%%%%%%%%%%%%%%Section 1 %%%%%%%%%%%%%%%%%%%%%%%%%%%%%%%%%%%%%%%%%%%%%%%%
\section{Introduction}
\label{section 1}

\paragraph{Bayesian networks.} 
Bayesian networks (BNs, for short) are one of the most prominent class of probabilistic graphical models \cite{DBLP:books/daglib/0023091} in AI.
They are used in very different domains, both for knowledge representation and reasoning. 
BNs represent conditional dependencies between random variables yielding -- if these dependencies are sparse -- compact representations of their joint probability distribution.
Probabilistic inference is the prime evaluation metric on BNs.
It amounts to compute conditional probabilities.
It is computationally hard: PP-complete~\cite{DBLP:journals/ai/Cooper90,DBLP:journals/ai/DagumL93}. 
A vast amount of inference algorithms exists, both exact ones (possibly tailored to specific graph structures such as bounded tree-width graphs), as well as advanced approximate and simulation algorithms.
State-of-the-art symbolic exact inference use different forms of decision diagrams.
In particular, sentential decision diagrams (SDDs for short~\cite{DBLP:conf/aaai/XueCD12}) and their probabilistic extension (PSDDs~\cite{DBLP:conf/kr/KisaBCD14}) belong to the prevailing techniques. 

\paragraph{Probabilistic model checking.}
Model checking of Markov chains and non-determ\-inis\-tic extensions thereof is a vibrant field of research since several decades.
The central problem is computing reachability probabilities, i.e., what is the probability to reach a goal state from a given start state?
Algorithms for computing conditional probabilities have been considered in~\cite{DBLP:conf/tacas/BaierKKM14}.
Efficient model-checking algorithms have been developed and tools such as PRISM~\cite{DBLP:conf/cav/KwiatkowskaNP11} and storm~\cite{DBLP:conf/cav/DehnertJK017} have been applied to case studies from several different application areas.
Like ordinary model checking, the state-space explosion problem is a major practical obstacle.
As for BNs, the use of decision diagrams has attracted a lot of attention since its first usages in probabilistic model checking~\cite{DBLP:conf/icalp/BaierCHKR97,DBLP:conf/tacas/AlfaroKNPS00} and improvements are still being developed, see e.g.,~\cite{DBLP:conf/tacas/0001BCDDKM016}.
As demonstrated in the QComp 2019 competition, MTBDD-based model checking prevails on various benchmarks~\cite{DBLP:conf/tacas/HahnHHKKKPQRS19}.

\paragraph{Topic of this paper.}
The aim of this work is to investigate to what extent off-the-shelf techniques from probabilistic model checking can be used for exact probabilistic inference in BNs.
We are in particular interested to study the usage of MTBDD-based symbolic model checking for inference, and to empirically compare its performance to inference using state-of-the-art decision diagrams in AI such as SDDs and their probabilistic extension.
To that end, we define a simple mapping from (discrete) BNs into discrete-time Markov chains (MCs) and relate Bayesian inference to computing reachability probabilities.
We report on an experimental evaluation on BNs of the \texttt{bnlearn} repository varying in size from small to huge (BN categorization) using a prototypical implementation on top of the storm model checker.
Our experiments show that inference using MTBDD-based model checking is quite sensitive to the evidence in the inference query, both in terms of size (i.e., the number of random variables) and depth (i.e., the ordering).
For BNs of small to large size, MTBDD-based symbolic model checking is  competitive to BN-specific symbolic techniques such as PSDDs whereas for very large and huge BNs, PSDD techniques prevail.

\paragraph{Contributions.} Our work aimed to reduce the gap between the area of probabilistic model checking and probabilistic inference. 
Its main contributions are:
\begin{itemize}
\item A simple mapping from Bayesian networks to Markov chains
\item A prototypical tool chain to enable model-checking based inference.
\item An experimental evaluation to compare off-the-shelf MTBDD-based inference by model checking to tailored PSDD inference.
\end{itemize}

\paragraph{Related work.}
There is a large body of related work on exploiting verification and/or symbolic data structures to inference. 
We here concentrate on the most relevant papers.
Deininger~\emph{et al.}~\cite{DBLP:conf/atva/DeiningerDM16} is perhaps the closest related work.
They apply PCTL model checking on factored representations of dynamic BNs and 
compare an MTBDD-based approach using partitioned representations of the transition probability matrix to monolithic representations.
Their experiments show that quantitative model checking does not significantly benefit from partitioned representations.
Langmead \emph{et al.} \cite{langmead2006temporal,langmead2008towards} employ probabilistic model checking algorithms to perform inference on Dynamic Bayesian Networks. They emphasize on broadening the queries, using temporal logic as the query language. 
Holtzen \emph{et al.}~\cite{DBLP:journals/corr/abs-1904-02079} consider symbolic inference on discrete probabilistic programs. 
They generalize efficient inference techniques for BNs that exploit the BN structure to such programs.
The key is to compile a program into a weighted Boolean formula, and to exploit BDD-based techniques to carry out weighted model counting on such formulas.
The works by Darwiche \emph{et al.} \cite{DBLP:conf/kr/Darwiche02,DBLP:journals/corr/abs-1301-3847,DBLP:conf/ecai/Darwiche04,DBLP:conf/ijcai/ChaviraD05} compile BNs into arithmetic circuits (via CNF) and perform inference by mainly differentiating these circuits in linear time in terms of the circuit sizes. This method is also applicable to relational BNs~\cite{DBLP:journals/ijar/ChaviraDJ06}. 
Minato \emph{et al.}~\cite{DBLP:conf/ijcai/MinatoSS07} propose an approach for exact inference by compiling BNs directly into multiple linear formulas using zero-suppressed BDDs.
This differs from Darwiche's approach as it does not require the intermediate compilation into CNF.
Shih \emph{et al.}~\cite{DBLP:conf/pgm/ShihCD18} propose a symbolic approach to compile BNs into reduced ordered BDDs in order to verify them against monotonicity and robustness.
Sanner and McAllester~\cite{DBLP:conf/ijcai/SannerM05} propose an affine version of algebraic decision diagrams to compactly represent context-specific, additive, and multiplicative structures. 
They proved that the time and memory footprint of these affine ADDs for inference can be linear in the number of variables in cases where ADDs are exponential.
Batz \emph{et al.}~\cite{DBLP:conf/esop/BatzKKM18} use deductive verification to analyse BNs. They show that BNs correspond to a simple form of probabilistic programs amenable to obtaining closed-form solutions for exact inference. 
They exploited this principle to determine the expected time to get one sample from the BN under rejection sampling. 
Approximate model checking has been applied to verify dynamic BNs against finite-horizon probabilistic linear temporal properties \cite{DBLP:conf/atva/PalaniappanT12}.
Finally, we mention the works \cite{DBLP:conf/aaai/ShachterDF90,DBLP:conf/cav/GehrMV16} that use exact symbolic inference methods, so-called PSI tools, on belief networks and probabilistic programs, basically through mapping program values to symbolic expressions.

\paragraph{Outline.} Section 2 introduces Bayesian networks and probabilistic inference. Section 3 briefly recapitulates Markov chain model checking. Section 4 presents the various symbolic data structures that are relevant to this work. Section 5 details how BNs are mapped onto Markov chains and how inference can be reduced to computing reachability probabilities. Section 6 reports on the experimental results, while Section 7 concludes the paper.

%%%%%%%%%%%%%%%%%%%%%%%%%%%%%%%%%%%%%%%%Section 2 %%%%%%%%%%%%%%%%%%%%%%%%%%%%%%%%%%%%%%%%%%%%%%%%%%%%%
\section{Bayesian Networks}
\label{section 2}
\begin{figure}[hbt!]
 	 \begin{center}
 	 \includegraphics[width=0.8\textwidth]{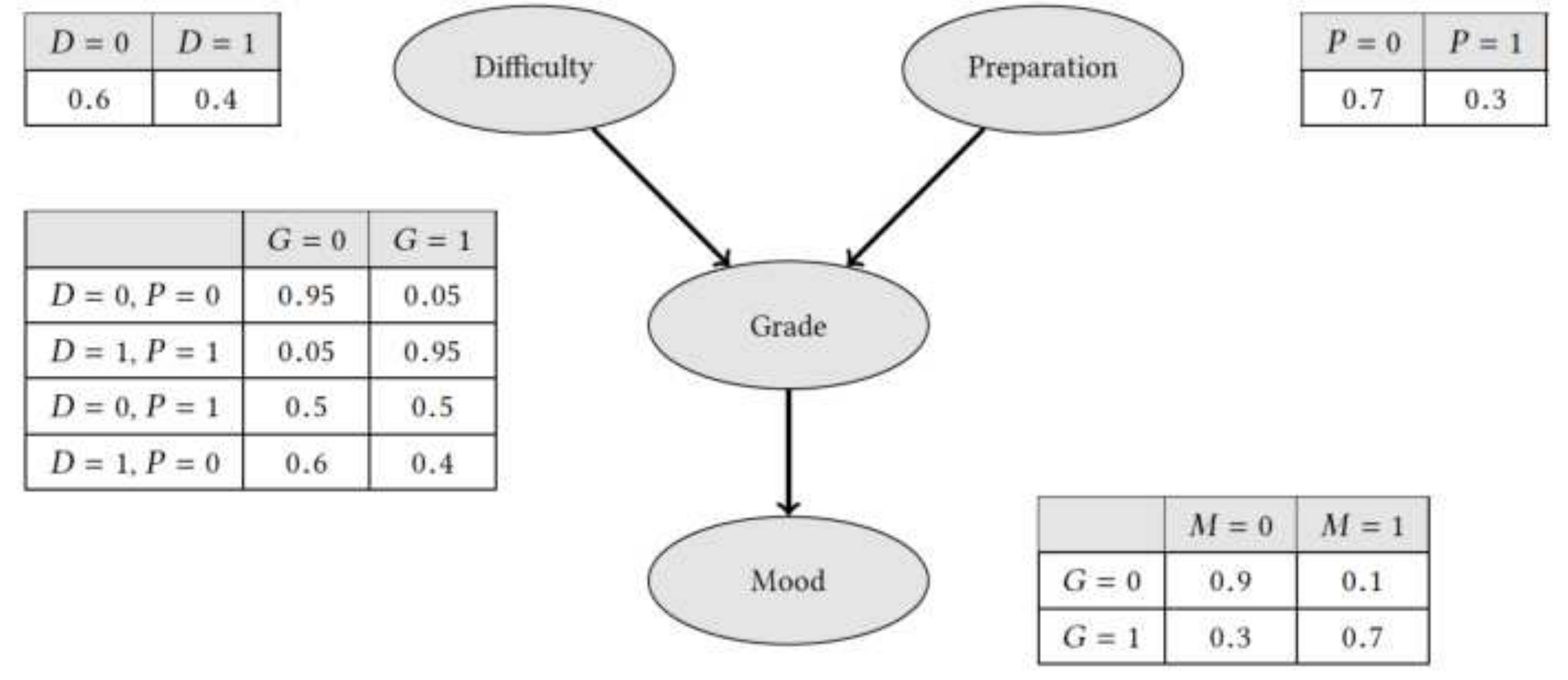}
 	\end{center}
 \caption{Simple example of Bayesain networks - Student Mood} \label{fig1}
\end{figure}

A {Bayesian network} (BN for short) is a tuple $B = (G, \Theta)$ where $G = (V, E)$ is a directed acyclic graph with finite set of vertices $V$ in which $v \in V$ represents a \emph{random variable} taking values from the finite domain $D$ and edge $(v, w) \in E$ represents the \emph{dependency} of $w$ on $v$. We let $parents(v) = \{ w \in V \mid (w, v) \in E\}$. For each vertex ${v}$ with ${k}$ parents, the function ${\Theta_v : D^k \to 	Dist(D)}$ is the \emph{conditional probability table} of (the random variable corresponding to) vertex ${v}$. $Dist(D)$ here denotes the set of probability distribution functions on $D$. Figure \ref{fig1} indicates a small BN, in which all the variables are binary. The DAG indicates the dependencies between the variables. For example, the grade a student gets for an exam depends on whether the exam has been easy or difficult, and additionally on whether she has been prepared for the exam. See Figure \ref{fig1}. \\ 
The conditional probability table $\Theta_v$ (CPT for short) of vertex $v$ defines a probability distribution which determines the evaluation of $v$, given some evaluation of $parents(v)$. For example, according to the CPT of Grade, the probability of a low grade is 0.95 for an easy exam and non-prepared student.

The semantics of BN $B = (V, E, \Theta)$ is the joint probability function that it defines. Let $W \subseteq V$ be a downward-closed set of vertices where $w \in W$ has value $\underline{w} \in D$. The unique joint probability function of BN $B$ equals:
\begin{equation}
Pr(W = \underline{W} ) = \prod_{w \in W} Pr(w = \underline{w} \mid parents(w) = \underline{parents(w)}) 
\end{equation}
In this paper, we are interested in probabilistic inference. Let $B$ be a BN with set $V$ of vertices, $F \subseteq V$ be the evidence, and $H \subseteq V$ be the hypothesis. The evidence can be simply seen as what we already know and the hypothesis as what we are interested in, given the evidence. The problem of \emph{(exact) probabilistic inference} is to determine the following conditional probability:
\begin{equation}
\label{eq 0}
	Pr(H = h \mid F = f) = \dfrac{Pr(H = h \wedge F = f)}{Pr(F = f)}
\end{equation}
In case $Pr(F = f) =0$, the query is considered ill-conditioned. In the student mood example, shown in Figure \ref{fig1}, let assume that we are interested to know how likely a student ends up with a bad mood after getting a bad grade for an easy exam, given that she is well prepared. This is defined as:
\begin{equation*}
\begin{split}
Pr(D = 0, G = 0, M = 0 \mid P = 1) = \dfrac{Pr(D = 0, G = 0, M = 0, P = 1)}{Pr(P = 1)} \\
= \dfrac{0.6 \cdot 0.5 \cdot 0.9 \cdot 0.3}{0.3} = \dfrac{0.081}{0.3} = 0.27
\end{split}
\end{equation*}
The decision variant of probabilistic inference can be defined for a given probability $p \in \mathbb{Q} \cap [0, 1)$ as follows:
\begin{center}
Does $Pr(H = h \mid F = f) > p$? 
\end{center}
This problem is PP-complete \cite{DBLP:books/daglib/0024906}. The \emph{average Markov blanket} of a BN is an indication of the practical complexity of performing inference. The Markov blanket for a vertex $v$ in a BN is the set $\partial_v$ composed of the parents, the children, and the spouses of $v$ \cite{DBLP:books/daglib/0024906}, where the spouses of $v$ are the nodes that have some common children with $v$. It follows that $ Pr(v \mid \partial_v \wedge w) = Pr(v \mid \partial_v)$, for any $w \in V$. The average Markov blanket of a BN, or AMB for short, then is the average size of the Markov blanket of all its vertices, that is, $\dfrac{1}{\lvert{V}\rvert}\sum_{v \in V} \lvert\partial_v\rvert$. AMB indicates the average degree of dependence between the random variables in the BN.

%%%%%%%%%%%%%%%%%%%%%%%%%%%%%%%%%%%%%%%%Section 2 %%%%%%%%%%%%%%%%%%%%%%%%%%%%%%%%%%%%%%%
\section{Markov Chain Model Checking}
\label{section 3}
Since in this work we are focused on discrete time BNs, we are interested in discrete-time Markov chains. DTMCs or simply MCs for short, are simple probabilistic models that equip each edge with a probability.
An MC $M$ is a tuple $(\Sigma, \sigma_I, P)$ where $\Sigma$ is a countable non-empty set of states, $\sigma_I$ is the initial state, and $P : S \rightarrow Dist(\Sigma)$ is the transition probability function. \\
In this work we are interested in computing reachability probabilities in an MC. The reachability probability for $G \subseteq \Sigma$ is defined as the probability of finally reaching $G$, starting from the initial state $\sigma_I$. This is denoted by $Pr_M(\lozenge G)$.
Computing $Pr_M(\lozenge G)$ can be reduced to computing the unique solution of a linear equation system \cite{DBLP:conf/lics/Katoen16} whose size is linear in $|\Sigma|$. This can be done in a symbolic manner using MTBDDs \cite{DBLP:conf/icalp/BaierCHKR97}.

%%%%%%%%%%%%%%%%%%%%%%%%%%%%%%%%%%%%%%%%Section 4 %%%%%%%%%%%%%%%%%%%%%%%%%%%%%%%%%%%%%%%%%%%%%%%%%
\section{Symbolic Data Structures}
\label{section 4}
The need to represent Boolean functions and probability distributions in a
succinct manner has led to various compact representations \cite{DBLP:reference/mc/Bryant18,DBLP:journals/fmsd/BaharFGHMPS97,DBLP:journals/fmsd/FujitaMY97,DBLP:conf/ijcai/Darwiche11,DBLP:conf/kr/KisaBCD14}. Symbolic model checking mainly relies on set-based and binary encoding of states and transitions enabling the use of compact representations such as BDDs and MTBDDs. In the following we briefly review the data structures related to this work: BDDs, MTBDDs, vtrees, SDDs and PSDDs. The first two are popular in symbolic model checking while the last three are state-of-the-art in probabilistic inference. %[reference].
\subsection{Reduced Ordered Binary Decision Diagrams}
ROBDDs or simply BDDs for short, are dominantly-used structures for representing switching functions. BDDs result from compacting binary decision trees mainly by eliminating \emph{don't care} nodes and duplicated subtrees. Essential characteristic of ROBDDs is that they are canonical for a given function and a given variable ordering \cite{DBLP:reference/mc/Bryant18}. Optimal variable orderings can yield very succinct ROBDDs. Although finding the optimal variable ordering is NP-hard \cite{DBLP:journals/tc/BolligW96}, ROBDDs can be very compact in practice \cite{DBLP:reference/mc/ChakiG18}.\\
Let $\wp=(z_1, ..., z_m)$ be a (total) variable ordering for $Var = \{z_1, ..., z_m\}$ where $z_1 <_\wp ... <_\wp z_m$. An $\wp\mhyphen OBDD$ is a tuple $\mathfrak{B} = (V, V_I, V_T, succ_0, succ_1, var, val, v_0)$ with the finite set $V$ of nodes, partitioned into inner nodes $V_I$ and terminal nodes $V_T$. $v_0 \in V_I$ is the unique distinguished root node. $succ_0, succ_1 : V_I \rightarrow V$ are the successor functions assigning a zero-successor $v_r \in V$ and a one-successor $v_l\in V$ to each $v \in V$. The labelling functions $var : V_I \rightarrow Var$ and $val : V_T \rightarrow \{0, 1 \}$ must satisfy the following equation for $v \in V_I$ and $w \in \{succ_0(v), succ_1(v)\}$:
\begin{equation} \label{equation 1}
(var(v) = z_i \land w \in V_I) \Rightarrow var(w) = z_j \quad with \quad z_i <_\wp z_j.
\end{equation} 
Every inner node $v$ in an OBDD represents a variable from $Var$. The terminal nodes are mapped to $0$ or $1$. Based on the evaluation of $var(v)$ either to $0$ or to $1$, the transition from $v$ to the next node is chosen from $\{succ_0(v), succ_1(v)\}$. 
The semantics of an $\wp \mhyphen OBDD$ is the switching function $f_\mathfrak{B}$ where $f_\mathfrak{B}([z_1 = b_1, ..., z_m = b_m])$ is determined by the value of the resulting leaf obtained by traversing the OBDD starting from the root $v_0$ and branching according to the evaluation $[z_1 = b_1, ..., z_m = b_m]$. \\
For terminal $v \in V_T$, $f_v$ represents the constant function $f_v$ with value $val(v)$. For $v \in V_I$, $f_v$ is defined based on the Shannon expansion over $v$ as $f_v = \left( \neg z \land f_{succ_0}(v)\right) \lor \left( z \land f_{succ_1}(v) \right)$, where $z = var(v)$. A $\wp \mhyphen OBDD$ $\mathfrak{B}$ is \emph{reduced} if for every pair $(v, w)$ of nodes in $\mathfrak{B}$, $v \neq w$ implies $f_v \neq f_w$. An OBDD can be reduced by recursively applying simple reduction rules: the elimination of \emph{don't care} vertices, and the elimination of isomorphic subtrees.

 \subsection{Multi-Terminal BDDs} While BDDs represent Boolean functions, the terminal values in MTBDDs can acquire values from other domains such as real or rational numbers. This allows rational or real functions to be succinctly represented, enabling representing probability distribution functions. The formal definition of MTBDD is not very different from the one for BDD. Let $Var$ and $\wp$ be as before. An MTBDD $\mathfrak{M}$ is the same structure as an OBDD except that (in our setting) the value function $val$ is refined to $val : V_T \rightarrow \left[0, 1\right]$ assigning each terminal node $v \in V_T$ a probability $val(v)$. The semantics of MTBDD $\mathfrak{M}$ is defined by $f_\mathfrak{M} : Eval(Var) \rightarrow \left[0, 1\right]$ similarly to $f_\mathfrak{B}$ for BDD $\mathfrak{B}$.
 
 \subsection{Sentential Decision Diagrams}
Sentential Decision Diagrams \cite{DBLP:conf/ijcai/Darwiche11} represent propositional knowledge bases. They are inspired by two concepts: \emph{structured decomposability} \cite{DBLP:conf/aaai/PipatsrisawatD08} which is based on vtrees, and the generalisation of Shannon decomposition which is \emph{strongly deterministic decomposition} \cite{DBLP:conf/aaai/PipatsrisawatD10}. 

\paragraph{vtree.} A vtree \cite{DBLP:conf/aaai/PipatsrisawatD08} for a set of variables $V$ is a
full (but not necessarily complete), rooted binary tree whose leaves represent the variables in $V$. The node $v$ and the subtree rooted at the node $v$ are often called the same. Let $var(v)$ indicate the set of variables stored in the leaves of the subtree rooted at the node $v$. Let $v^l$ and $v^r$ be respectively indicate the left and right children of $v$.
A Boolean function $f$ in Decomposable Negation Normal Form (DNNF) is said to \emph{respect} a vtree $T$ if for every conjunction $\alpha \land \beta$ in ${f}$, there is a node $t$ in $T$ such that $var(\alpha) \subseteq var(t^l)$ and $var(\beta) \subseteq var(t^r)$.

\paragraph{Strongly deterministic decomposition.} Let $f$ be a Boolean function with disjoint sets of variables $X$ and $Y$. If $f$ can be written as $(p_1(X) \land s_1(Y)) \lor ... \lor (p_n(X) \land s_n(Y))$, then $\{(p_1, s_1),...,(p_n,s_n) \}$ is called an $(X, Y)$-decomposition of $f$ in terms of Boolean functions $p_i$ and $s_i$ on $X$ and $Y$ respectively. Provided that $p_i \land p_j = false$ for $i \neq j$, the decomposition is called \emph{strongly deterministic} on $X$. Here, each $p_i$ is called a prime and each $s_i$ a sub. Let $\mathfrak{S}$ be a strongly deterministic $(X, Y)$-decomposition of function $f$. $\mathfrak{S}$ is called an $X$-partition of $f$ iff its primes make a \emph{partition}. This means each prime is consistent (i.e., can be true at some evaluation), every pair of distinct primes are mutually exclusive, and the disjunction of all primes is valid (true).

\paragraph{SDD.} An SDD can be seen as a recursive $(X, Y)$-strongly deterministic decomposition of a switching function according to a particular vtree, starting from the root node. The semantics of SDD $\mathfrak{S}$ is defined by the switching function $f_{\mathfrak{S}}$ with respect to the vtree $v$. $\mathfrak{S}$ is an SDD respecting vtree $v$ iff (1) $\mathfrak{S} = \bot$ or $\mathfrak{S} = \top$, with the semantics $f_{\bot} = false$ and $f_{\top} = true$, (2) $\mathfrak{S} = X$ or $\mathfrak{S} = \neg X$ and $v$ is a leaf with variable $X$ with the semantics $f_{X} = X$ and $f_{\neg X} = \neg X$, (3) $\mathfrak{S} = \{(p_1, s_1), ..., (p_n, s_n)\}$, $v$ is internal, $p_1, ..., p_n$ are SDDs that respect subtrees of $v^l$, and $s_1, ..., s_n$ are SDDs that respect subtrees of $v^r$, where $f_{p_1}, ..., f_{p_n}$ makes a partition. 
In this case, the semantics of SDD $\mathfrak{S}$ is given by $f_{\mathfrak{S}} = \bigvee\limits_{i=1}^n (p_i \land s_i)$.
 \\
An SDD is \emph{compressed} iff all its subs are distinct ($s_i \neq s_j$ for $i \neq j)$. Moreover, it is called \emph{trimmed} if it does not contain any decomposition of the form $\{(\top, \mathfrak{S})\}$ or $\{(\mathfrak{S}, \top), (\neg \mathfrak{S}, \bot)\}$. These two properties characterise the canonicity of SDDs as follows. Two SDDs that are compressed and trimmed and respect the same vtree are semantically equivalent if and only if they are equal \cite{DBLP:conf/ijcai/Darwiche11}. Thus, vtrees for SDDs resemble the role of variable ordering for BDDs. In the same manner, compression and trimming resemble reduction in BDDs.

\begin{figure}[hbt!]
  \hspace*{0.3cm}
  \begin{minipage}[b]{0.2\textwidth}
    \includegraphics[scale=0.8]{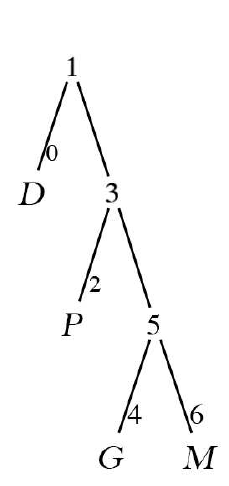}
  \end{minipage}
  \begin{minipage}[b]{0.65\textwidth}
    \includegraphics[scale=0.35]{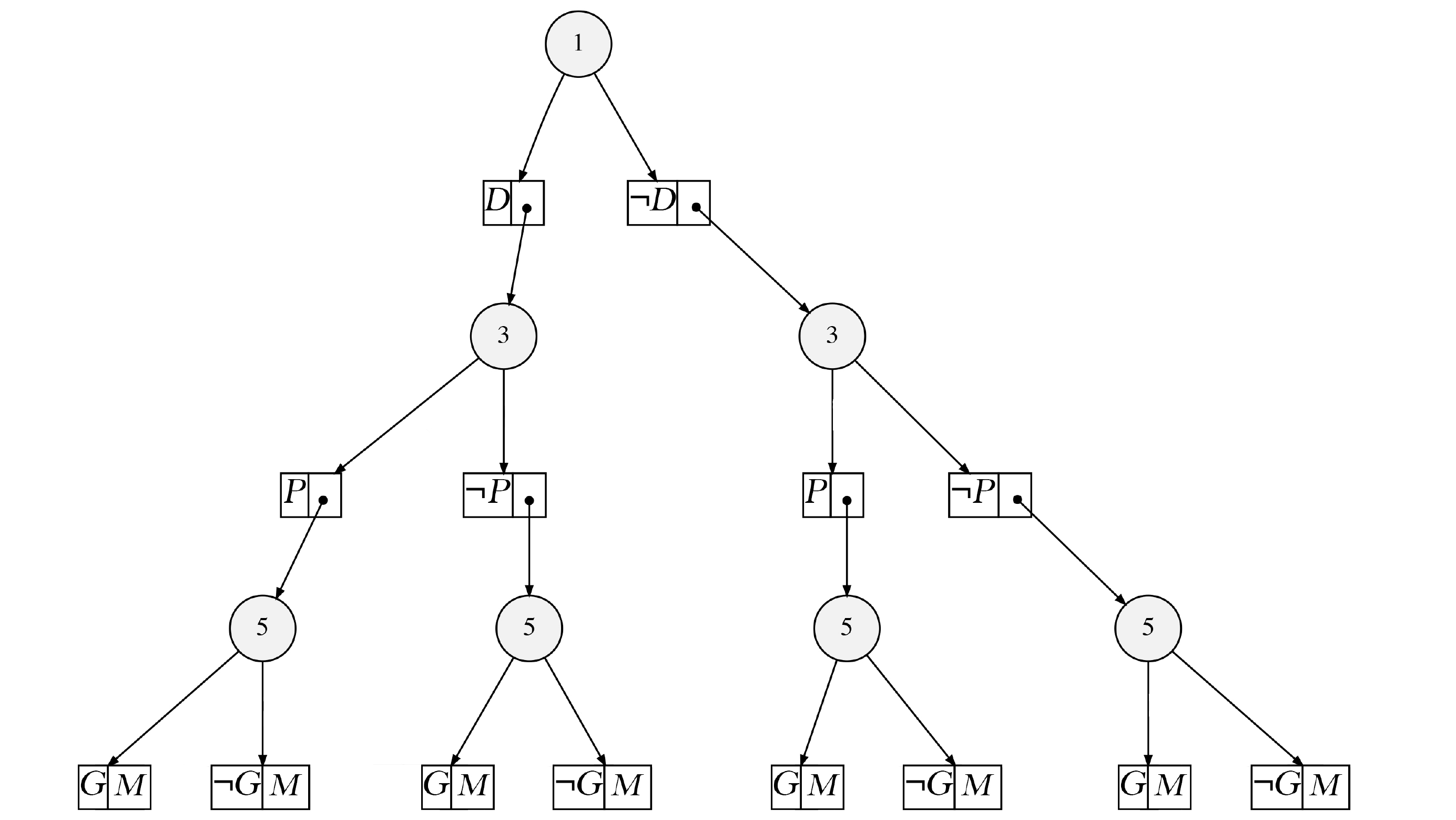}
  \end{minipage}
  \caption{An SDD (right) and the corresponding vtree (left) for Student Mood example } \label{fig: 33}
 \end{figure}
 
\begin{example}
\label{example:1}
Figure \ref{fig: 33} (right) indicates an SDD for the Student Mood example (see Figure \ref{fig1}). The underlying vtree is depicted in Figure \ref{fig: 33} (left). The two-parts boxes in SDD visually indicate the prime-sub pairs. The circles indicate decision nodes. Decision node $(p_1, s_1), ..., (p_k, s_k)$ has $k$ outgoing edges and edge $i$ is connected to $(p_i, s_i)$. The SDD respects the vtree in the sense that on each vtree node the leaves in the left subtree determine the primes and the leaves in the right subtree determine the subs. For instance, for node 3 in the vtree, $P$ determines the primes and $G$ and $M$ determine the subs. Since each decision node makes an $(X, Y)$-partition, each variable evaluation holds on exactly one prime.
\end{example}

\subsection{Probabilistic SDDs}
PSDDs \cite{DBLP:conf/kr/KisaBCD14} are recent representations in the domain of reasoning and learning. Similarly to MTBDDs, which are BDDs to represent non-Boolean functions, SDDs are extended to PSDDs in order to particularly represent probability distributions. A single SDD can be parameterized in infinitely many ways, each yielding a probability distribution. This is similar to BNs in a sense that each DAG can be extended to infinitely many $\Theta$s, (i.e., conditional probability tables) where each $\Theta$ specifies a probability distribution. PSDDs are complete in the sense that every distribution can be induced by a PSDD. PSDDs are canonical in the sense that for a given vtree, there is a unique trimmed and compressed PSDD. Interestingly, computing the probability of a term can be done in time linear in the PSDD size \cite{DBLP:conf/kr/KisaBCD14}.
\paragraph{Syntax.} A PSDD parametrizes an SDD in the following manner: (1) Every decision node $(p_i, s_i), ..., (p_k, s_k)$ and every prime $p_i$ is equipped with a positive parameter $\theta_i$ such that $\theta_1 + ... + \theta_k = 1$ and $\theta_i = 0$ iff $s_i = \bot$. The PSDD decision node is indicated by $(p_1, s_1, \theta_1), ..., (p_k, s_k, \theta_k)$. (2) For each terminal node $\top$, a positive parameter $\theta$ is supplied such that $0 < \theta < 1$. Syntactically, the terminal node $\top$ with parameter $\theta$ is indicated by $x:\theta$, where $x$ is the variable of the vtree leaf node that $\top$ is normalized \cite{DBLP:conf/ijcai/Darwiche11} for. Other terminal nodes ($\bot$, $x$, and $\neg x$) have fixed pre-defined parameters.

\paragraph{Semantics.} Let $n$ be a PSDD node respecting a vtree node $v$. Node $n$ represents the probability distribution $Pr_n$ over the variables of vtree $v$ defined by:
\begin{itemize}
    \item If $n$ is a terminal node and $v$ consists of variable $x$, then
    \begin{itemize}
        \item for $n = x: \theta$, $Pr_n(x) = \theta$ and $Pr_n(\neg x) = 1 - \theta$
        \item for $n = \bot$, $Pr_n(x) = 0$ and $Pr_n(\neg x) = 0$
        \item for $n = x$, $Pr_n(x) = 1$ and $Pr_n(\neg x) = 0$
        \item for $n = \neg x$, $Pr_n(x) = 0$ and $Pr_n(\neg x) = 1$.
    \end{itemize} 
    \item If $n$ is a decision node $(p_1, s_1, \theta_1), ..., (p_k, s_k, \theta_k)$ and $v$, the corresponding vtree node, has $X$ as left variables and $Y$ as right variables,
        $Pr_n(\underline{X}, \underline{Y}) = Pr_{p_i}(\underline{X}) \cdot  Pr_{s_i}(\underline{Y}) \cdot \theta_i$ for $i$ that $\underline{X} \models p_i$.
        Here, $\underline{X}$ denotes the evaluation of $X$'s variables. The condition $\underline{X} \models p_i$ holds on exactly one of the primes, by definition of $(X, Y)$-decomposition.
\end{itemize}

\begin{example}
Figure \ref{fig: 44} (right) denotes the PSDD for the Student Mood example, with the same underlying vtree as in Example \ref{example:1}. The visual representation of PSDD extends SDD in two manners: (1) edge $i$ directing from the decision node $(p_1, s_1, \theta_1), ..., (p_k, s_k, \theta_k)$ to the pair $(p_i, s_i)$ is labeled with $\theta_i$, (2) terminal node $\top$ is replaced by $x: \theta$, according to the PSDD syntax and semantics. The PSDD  in Figure \ref{fig: 44} (right) induces the same probability distribution function induced by the Student Mood BN. For instance, $Pr(\neg D \wedge P \wedge \neg G \wedge \neg M) = 0.6 \cdot  0.3 \cdot( 0.5 \cdot (1 - 0.1)) = 0.081$, by the PSDD semantics.

\begin{figure}[hbt!]
  \hspace*{0.3cm}
  \begin{minipage}[b]{0.15\textwidth}
    \includegraphics[scale=0.8]{images/studentMoodVtree.pdf}
  \end{minipage}
  \begin{minipage}[b]{0.7\textwidth}
    \includegraphics[scale=0.35]{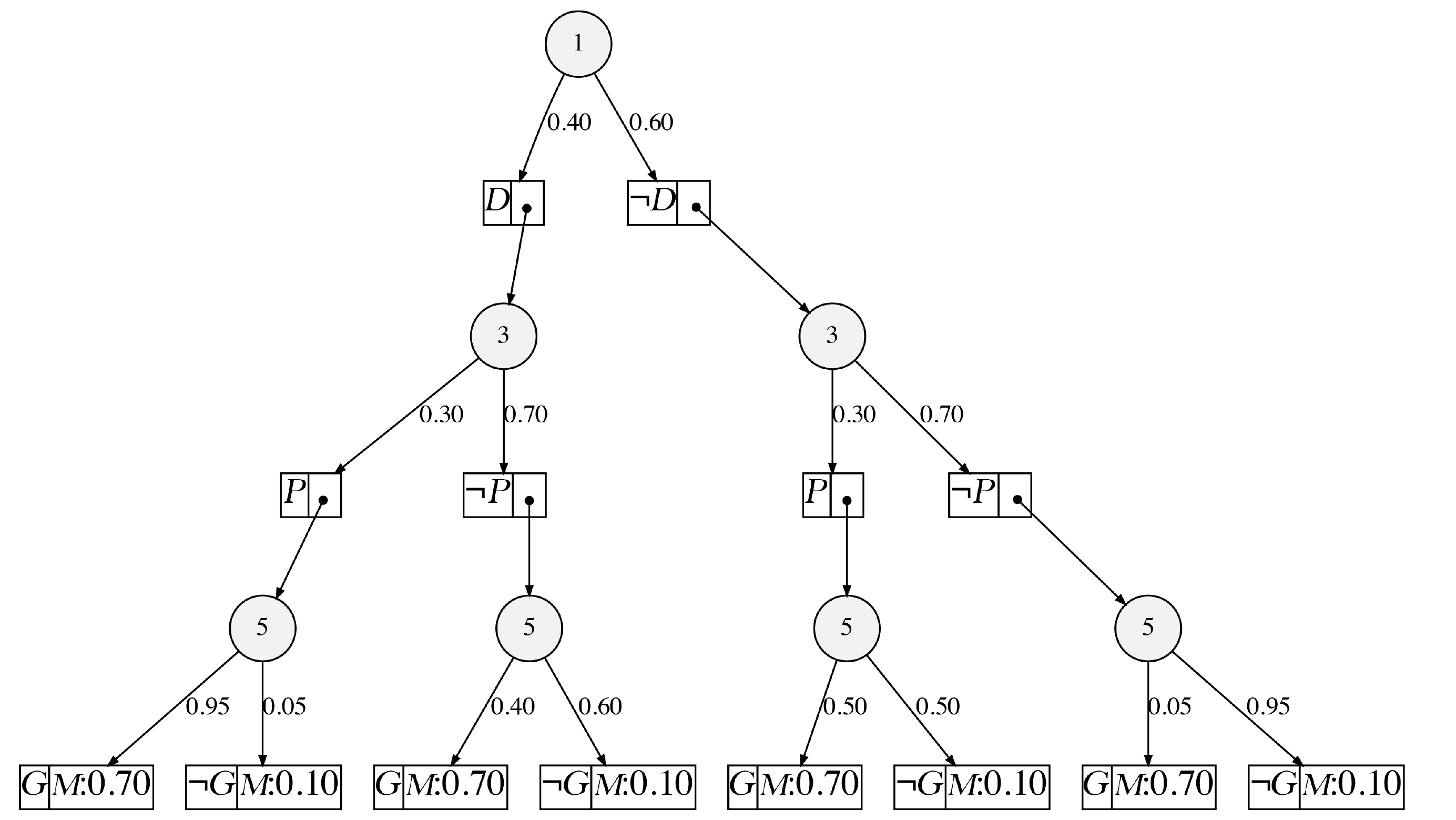}
  \end{minipage}
  \caption{A PSDD (right) and vtree (left) compiled from the Student Mood example } \label{fig: 44}
\end{figure}
\end{example}

%%%%%%%%%%%%%%%%%%%%%%%%%%%%%%%%%%% Section 5 %%%%%%%%%%%%%%%%%%%%%%%%%%%%%%%%%%%%%%%%%%%%%%
\section{BN Analysis using Probabilistic Model Checking}
\label{section 5}
In this section, we are going to explain our approach in detail. First, let us explain some notations. Let $X$ be a set of variables. Let $\underline{X}$ denote the evaluation of $X$'s variables. We use $*$ to denote a don't care value. We use $\mu$ to denote a probability distribution. Let $\prod_{i = 1}^n X_i = X_1 \times ... \times X_n = \{ (x_1, ..., x_n) | x_1 \in X_1 , ..., x_n \in X_n \}$ be the Cartesian product over the sets $X_i, ..., X_n$. \\
The basic idea for the transformation is to map a Bayesian network $B$ onto the Markov chain $M_B$ such that the conditional reachability probabilities in $M_B$ correspond to the conditional probabilistic inference queries in $B$. Colloquially stated:
\begin{equation}
\label{eq1}
	Pr_B(H = h \mid F = f) = Pr_{M_B}(\lozenge(H = h) \mid \lozenge(F = f)).
\end{equation}
The definition of MC $M_B$ is as follows.
\begin{definition}{\textbf{(The Markov chain of a BN).} } \label{Def: MCBDef}
Let $B = (V, E, \Theta) $ be a BN with $V = \{v_1, ..., v_n\}$ and $dom(v_i) = D_i$ with elements $d_i \in D_i$. For $\varrho = (v_1, ..., v_n)$ a \emph{topological order} on the DAG $(V, E)$, let MC $M_B = (\Sigma, \sigma_I, P)$ be the Markov chain of $B$ where:
\begin{itemize}
\item $\Sigma = \displaystyle{\prod_{i = 1}^n \{v_i\} \times (D_i \cup \{*\} )}$ is the set of \emph{states},
\item $\sigma_I = V \times \{*\}$ is the \emph{initial state}, and
\item $ P : \Sigma \times \Sigma \rightarrow [0, 1] $ is the \emph{transition probability function} defined by the following SOS rules:
\end{itemize}

\begin{equation}
\label{eq2} 
\begin{aligned}
\dfrac{\Theta_{v_1} = \mu, \quad  \mu(d_1) = p}{\sigma_I \xrightarrow{p} \sigma_{(v_1, d_1)} = {((v_1, d_1),(v_2, *), ..., (v_n, *))}}
\end{aligned}
\end{equation}

\begin{align}
\label{eq3}
\begin{split}
\dfrac{\splitdfrac{\quad \Theta_{v_i}(\underline{parents(v_i)}) = \mu, \quad \mu(d_i) = p,}
    {parents(v_i) \times \underline{parents(v_i)} \subseteq \{(v_1, d_1), ..., (v_{i-1}, d_{i-1})\}}}
 {\splitdfrac{\sigma = ((v_1, d_1), ... , (v_{i-1}, d_{i-1}), (v_i, *), ..., (v_n, *))}
 {\xrightarrow{p} \sigma' = ((v_1, d_1), ... , (v_i, d_i),(v_{i+1}, *), ..., (v_n, *))}}
\end{split}
\end{align}

\begin{equation}
\label{eq4}
\begin{aligned}
{\sigma = ((v_1, d_1), ..., (v_n, d_n))\xrightarrow{1}\sigma = ({(v_1, d_1), ..., (v_n, d_n))}}.
\end{aligned}
\end{equation} 
\end{definition}

The states in MC $M_B$ are tuples of pairs in the form of $(v, d)$ where $v$ is a variable of BN $B$ and $d \in dom(v) \cup \{*\}$ is the current value of $v$. The symbol $*$ is used to denote the initial evaluation of a variable. The initial state is $((v_1, *), ..., (v_n, *))$. The transition probability function specifies the probability of evolving between states. These transition probabilities correspond to the values in the conditional probability tables of $B$'s variables. The rule (\ref{eq2}) defines the transitions from the initial state to its successors according to $\Theta_{v_1}$. Since $v_1$ is the first variable in the topological order, $parents(v_1) = \emptyset$. If $\Theta_{v_1} = \mu$ and $\mu(d_1) = p$, there is a transition with probability $p$ from the initial state $\sigma_I$ to the state $\sigma$ in which all the variables are $*$ except for $v_1$ which is mapped to $d_1$.
Let states in which all the variables have taken values from their domain constitute the final states, $\{(v, d) \mid v \in V, d \neq *\}$. According to the rule (\ref{eq4}), the final states are equipped with a 1-probability self-loop.
The transitions from the states that are neither initial nor final are formalized in the rule (\ref{eq3}). Let $\underline{parents(v_i)}$ be the evaluation of all the variables in the set $parents(v_i)$. Let $\mu = \Theta(\underline{parents(v_i)})$. If $\mu(d_i) = p$, then from the state $\sigma$ where all variables before $v_i$ based on the ordering $\varrho$ are evaluated to values other than $*$, the transition goes to the state  $\sigma'$ where all the variable evaluations remain the same, except for $(v_i, *)$ that changes to $(v_i, d_i)$. 
The transition can take place only provided that all the variables in $parents(v_i)$ are already evaluated at the state $\sigma$ and their evaluation is consistent with the values in $\underline{parents(v_i)}$. This is ensured by the premise $parents(v_i) \times \underline{parents(v_i)} \subseteq \{ (v_1, d_1), ..., (v_{i-1}, d_{i-1})\}$. 

\begin{example}
Reconsider the BN Student Mood shown in the Figure \ref{fig1}. Figure \ref{fig2} (left) indicates the MC $M_{StudentMood}$ resulting from the above definition. Here, the \emph{don't care} evaluations are omitted from the states and the states related to "Mood" variable are ignored. The probabilities on evolving edges are based on their corresponding values in the conditional probability tables. For instance, for the left most path in the MC this is $0.4,\,0.3,$ and $0.95$. The last number is set, for example, according to the Grade's CPT where the probability of $Grade = 1$ equals $0.95$, for $Dif = 1$ and $Prep = 1$.
The MTBDD corresponding to the probability distribution over the leaves of $M_{StudentMood}$ is shown in Figure \ref{fig2} (right), again abstracting from \emph{don't care} values and "Mood" variable. The nodes binary evaluations are coded as the right and left edges in an MTBDD. For example, the left most path indicates all the variables being evaluated to $1$. The terminal nodes denote the joint probability, which in this case equals $0.4 \cdot 0.3 \cdot 0.95 = 0.114$.

\begin{figure}[hbt!]
  \begin{minipage}[b]{0.5\textwidth}
	 \includegraphics[scale=0.18]{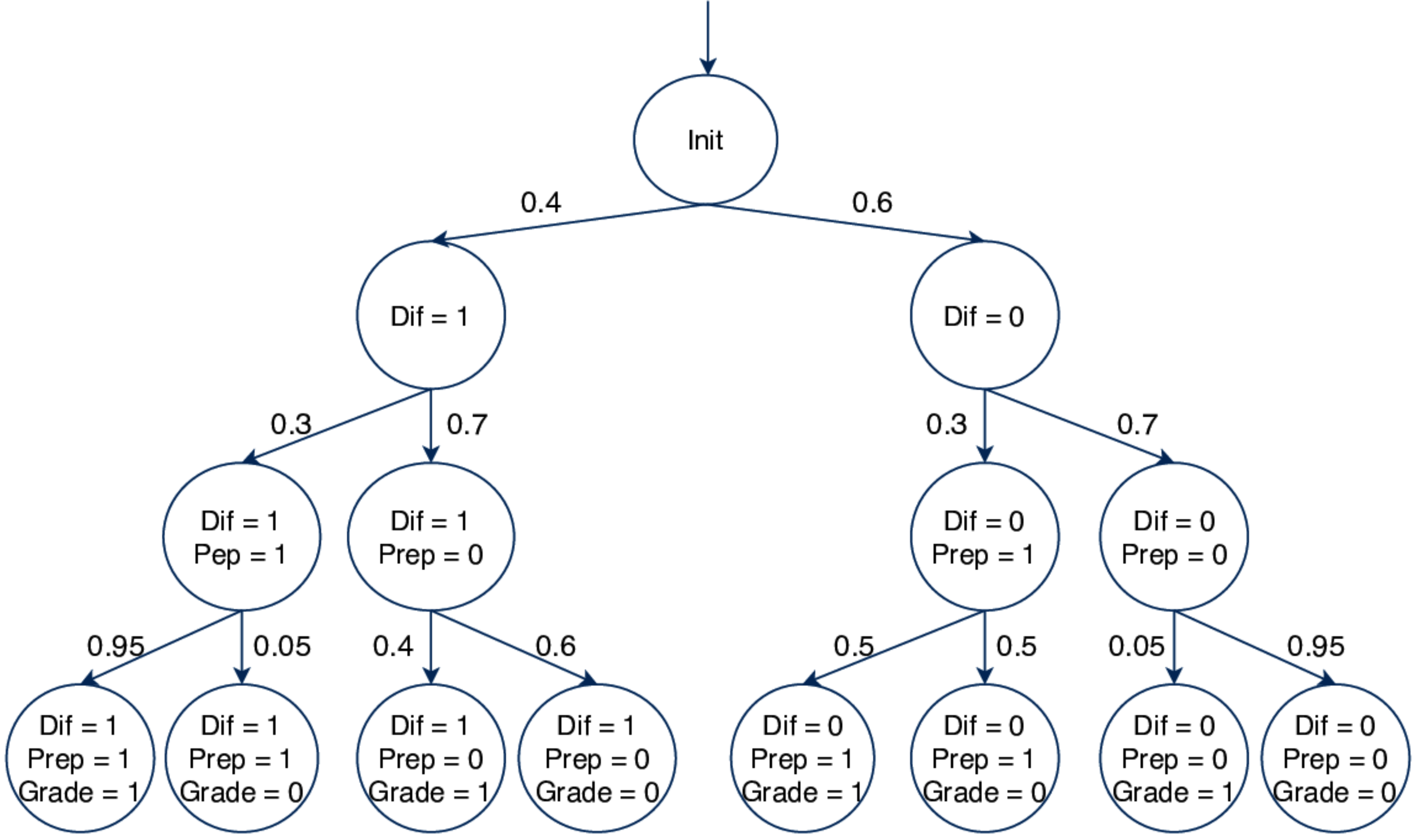}
  \end{minipage}
  \hfill
  \begin{minipage}[b]{0.4\textwidth}
    \includegraphics[scale=0.18]{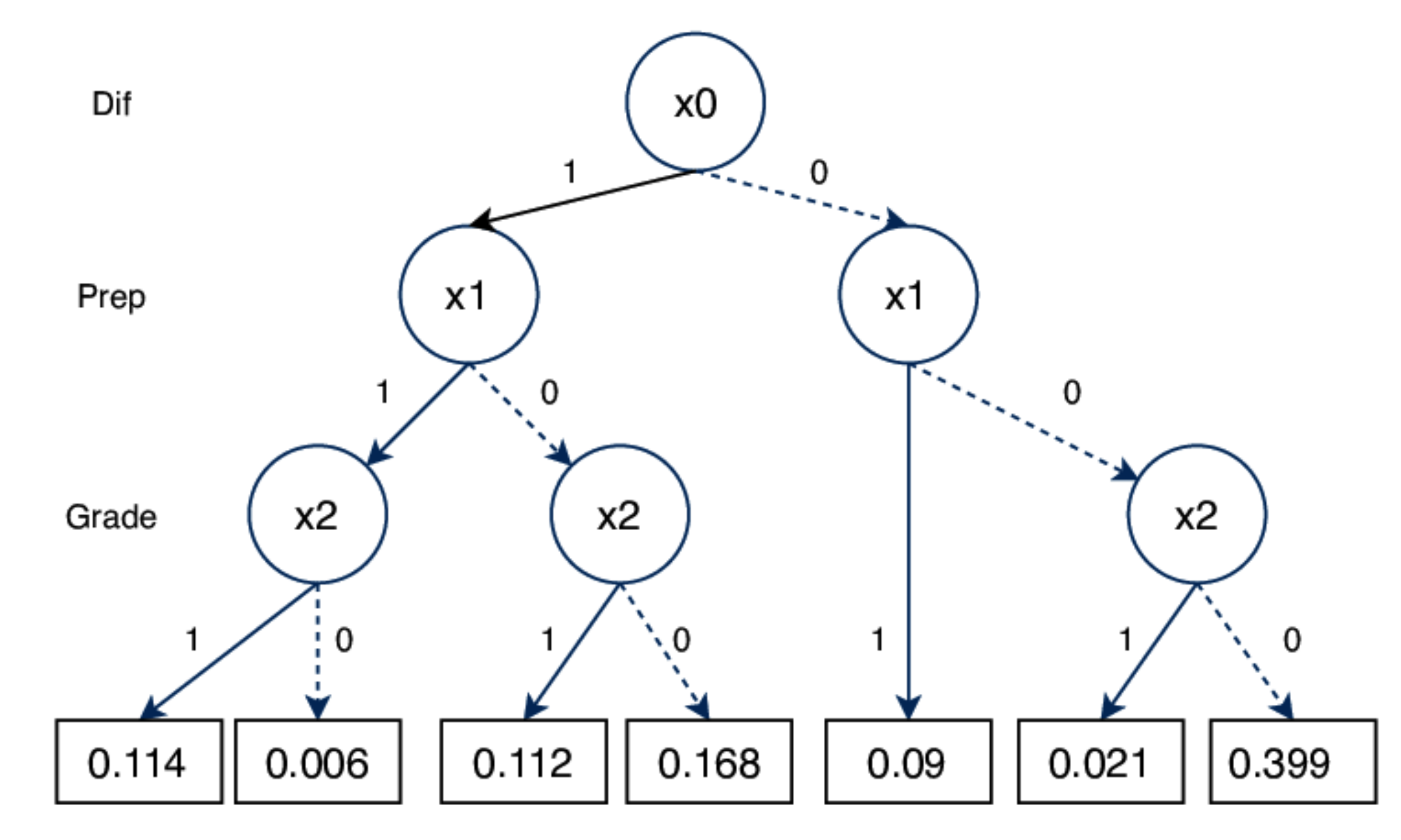}
  \end{minipage}
  \caption{The corresponding MC (left) and  MTBDD (right) for Student Mood example} \label{fig2}
\end{figure}	
\end{example}

\begin{proposition}{\textbf{(The size of MC $M_B$)}} \label{prop: 1}
Let $B$ be a BN with $dom(v_i) = D_i$ for each $v_i \in V$ and $\lvert M_B \rvert$ be the number of states in the Markov chain $M_B$. Then,
\begin{equation}
\label{eqsize}
	\lvert M_B \rvert \leq 1 + \sum_{i=1}^{n} \prod_{k=1}^{i} \lvert D_k \rvert \bigr ..
\end{equation}

\end{proposition}
In the special case where all random variables in $B$ have domain $D$
\begin{equation*}
\label{eqsize2}
\begin{split}
    |M_B| \leq  \sum_{i=0}^{n} \lvert D \rvert^{i},\, thus \quad |M_B| \leq \dfrac{1 - \lvert D \rvert^{\, n + 1}}{1 - \lvert D \rvert} .
    \end{split}
\end{equation*}
\begin{example}
The number of states in Figure \ref{fig2} (right) is $|M_{B}| = 2^4 - 1 = 15$.
\end{example}

We now consider the reachability probability as shown in the equation (\ref{eq1}). By definition of conditional probabilities we have:
\begin{equation}
\label{eq5}
		 Pr_{M_B} ( \lozenge(H = h) \mid   \lozenge(F = f)) = \dfrac{Pr_{M_B}( \lozenge(H = h) \wedge \lozenge(F = f)))}{Pr_{M_B}( \lozenge(F = f))}
\end{equation}

To determine the right-hand side we observe that given the tree structure of MC $M_B$ it holds:
\begin{proposition} \label{prop: 2}
For Markov Chain $M_B$ of BN $B$:
\begin{center}
    $Pr_{M_B}(\lozenge(H = h) \wedge \lozenge(F = f)) = Pr_{M_B}(\lozenge(H = h \wedge F = f))$.
\end{center}
\end{proposition}
From this, it is concluded that inference in BN $B$ can be reduced to computing reachability probabilities in MC $M_B$.

%%%%%%%%%%%%%%%%%%%%%%%%%%%%%%%%%%%%%%%%%% Section 6 %%%%%%%%%%%%%%%%%%%%%%%%%%%%%%%%%%%%%%%%%%%%%%%%%%%%
\section{Experimental Results}
\label{section 6}
\paragraph{Experimental setup.}We implemented a prototypical software tool for performing Bayesian inference as an extension of the probabilistic model checker Storm \cite{DBLP:conf/cav/DehnertJK017}. Our tool takes as input a BN in the Bayesian network Interchange Format \cite{bif} (BIF, for short). The BN is translated into a Markov chain as described in Definition \ref{Def: MCBDef}. The MC is specified using the Jani \cite{DBLP:conf/tacas/BuddeDHHJT17} modelling language, a high-level modelling language in the domain of probabilistic model checking. We evaluated our tool using various BN benchmarks from the Bayesian network repository \texttt{bnlearn} \cite{bnlearnrepository} that contains several BNs categorized in small, medium, large, very large, and huge networks. Table \ref{tab1} indicates some statistics of the evaluated BNs from the repository. The first column denotes whether all the variables in the BN are binary. The other statistics are the number of vertices, the number of edges, the maximum in-degree, the maximum domain size of variables, the average Markov blanket, and the number of parameters. The number of parameters is related to the total number of probabilities in all the conditional probability tables.
All our experiments were conducted on a $2.3$ GHz Intel Core i5 processor with 16 GB of RAM. \\
We focused our experimental validations on the following three questions:
\begin{enumerate}
    \item \label{Question: 1} What is the performance of MTBDD-based symbolic probabilistic model checking on Bayesian inference?
    \item \label{Question: 2} What is the effect of the number of observations and their depth in the topological ordering on the inference time in our approach?
    \item \label{Question: 3} How does inference using MTBDDs compares to PSDD techniques in terms of compilation time and inference time?
\end{enumerate}

\begin{table}
\centering
\caption{Statistics on the evaluated Bayesian networks in \texttt{bnlearn}} \label{tab1}
    \begin{tabular}{|c|l|l|l|l|l|l|l|l|}
        \hline
        {Binary} & {BN} & {\#Vertices} & {\#Edges}  & {InDegreeMax} & {Dmax}  & {AMB} & {\#Parameters}\\ 
        \hline \hline
        \multirow{5}{*}{YES} & cancer & 5 & 4 & 2 & 2 & 2.00 & 10 \\ 
        & earthquake & 5 & 4 & 2 & 2 & 2.00 & 10  \\
        & asia & 8 & 8 & 2 & 2 & 2.5 & 18  \\ 
       	& win95pts & 76 & 112 & 7 & 2  & 5.92 & 574\\
       	& andes & 223 & 338 & 6 & 2 & 5.61 & 1157  \\
       	\hline
        \multirow{9}{*}{NO} & survey & 6 & 6 & 2 & 3 & 2.67 & 21 \\
      	& sachs & 11 & 17 & 3 & 3 & 3.09 & 178 \\
      	& child & 20 & 25 & 2 & 6 & 3.00 & 230 \\
       	& alarm & 37 & 46 & 4 & 4 & 3.51 & 509 \\
       	& insurance & 27 & 52 & 3 & 5 & 5.19 & 984\\
       	& hepar2 & 70 & 123 & 6 & 4 & 4.51 & 1453 \\
       	& hailfinder & 56 & 66 & 4 & 11  & 3.54 & 2656 \\
       	& water & 32 & 66 & 5 & 4  & 7.69 & 10083\\ 
        & pathfinder & 135 & 200 & 5 & 63 &3.81 & 72079  \\
     \hline
    \end{tabular}
\end{table}

\paragraph{Bayesian inference using MTBDD-based model checking.}
In order to answer the first question, we have fed the Jani descriptions of the MCs into storm's sparse engine, and storm's bdd engine. The former fully builds a sparse matrix of the Markov chain $M_B$ of BN $B$, while the latter generates an MTBDD representation from the Jani file. The variable ordering of the MTBDD is determined based on the topological order of the BN. Table \ref{tab2} indicates the size of the resulting data structures and the compilation time. Here $E19$, for instance, denotes $10^{19}$ as the order of magnitude. The inference time on the sparse representation is prohibitive, even for medium-sized BNs, while inference using MTBDDs is mostly a matter of a few seconds or less. The large MC sizes are due to the exponential growth of the state space in the domain of the BN's variables, see Proposition \ref{prop: 1}. The significant size reduction with MTBDDs is due to the symmetrical and repetitive structure of the $M_B$. Those kind of symmetries and duplicated subtrees are merged in the MTBDD representation. The type of MTBDD shown in Figure \ref{fig2} (right) represents a discrete probability distribution. However, the MTBDD generated by storm encodes the Markov chain, i.e. its terminal nodes carry the transition probabilities of the Markov chain. This makes sharing of the subgraphs much more likely.

\begin{table}
\centering
\caption{Analysis of BN benchmarks using storm symbolic engine compared to sparse}\label{tab2}
    \begin{tabular}{|c|l| l l | l l|} 
        \hline
        &
        &
        \multicolumn{2}{l}{$Size$} &
        \multicolumn{2}{l|}{$Construction \: Time$} \\
        \hline
         Binary & BN & MC (\#states) & MTBDD (\#nodes) & Sparse Engine & Symbolic Engine \\ 
        \hline \hline
        % studentmood & 31 & 40 & 0.019s &  0.006s\\ \hline
        \multirow{5}{*}{YES} & cancer & 63 & 56 & 0.018s &  0.007s\\
        & earthquake & 63 & 55 & 0.023s & 0.006s \\
        & asia & 278 & 154 & 0.028s &  0.011s\\
        & win95pts & E19 & 446752 & $>$1.5h  & 11s \\
        & andes & E67 & 339485 & $>$1.5h  & 180s \\
        \hline 
        \multirow{9}{*}{NO} & survey & 238 & 70 & 0.031s & 0.008\\
        & sachs & 265720 & 165 & 0.469s &  0.072s\\
        & child & E9 & 731 & $>$1.5h & 0.277s\\
        & alarm & E16 & 2361 & $>$1.5h & 1s \\
        & insurance & E11 & 102903 &$>$1.5h & 2s\\ 
        & hepar2 & E42 & 7675 & $>$1.5h & 17s \\
        & hailfinder & E17 & 152201 & $>$1.5h  & 18s \\
        & water & E9 & 64744 & $>$1.5h & 20s\\
        & pathfinder & E242 & MO  & $>$1.5h  & -\\ [1ex] 
        \hline
\end{tabular}
\end{table}

\begin{figure}[hbt!]
\hspace*{-0.8cm}
  \begin{minipage}[b]{0.45\textwidth}
    \includegraphics[scale=0.45]{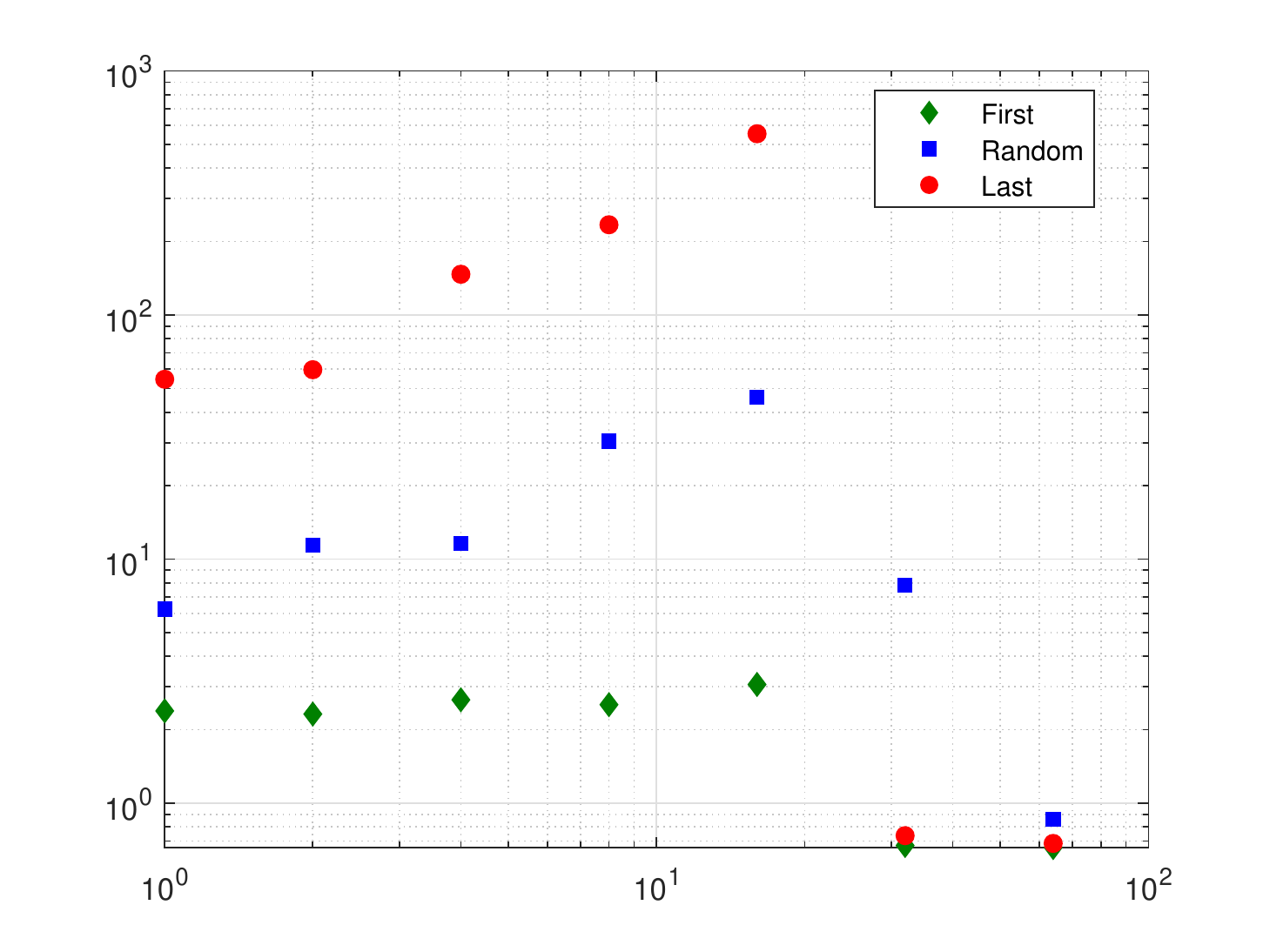}
  \end{minipage}
  \hspace{1cm}
  \begin{minipage}[b]{0.45\textwidth}
    \includegraphics[scale=0.45]{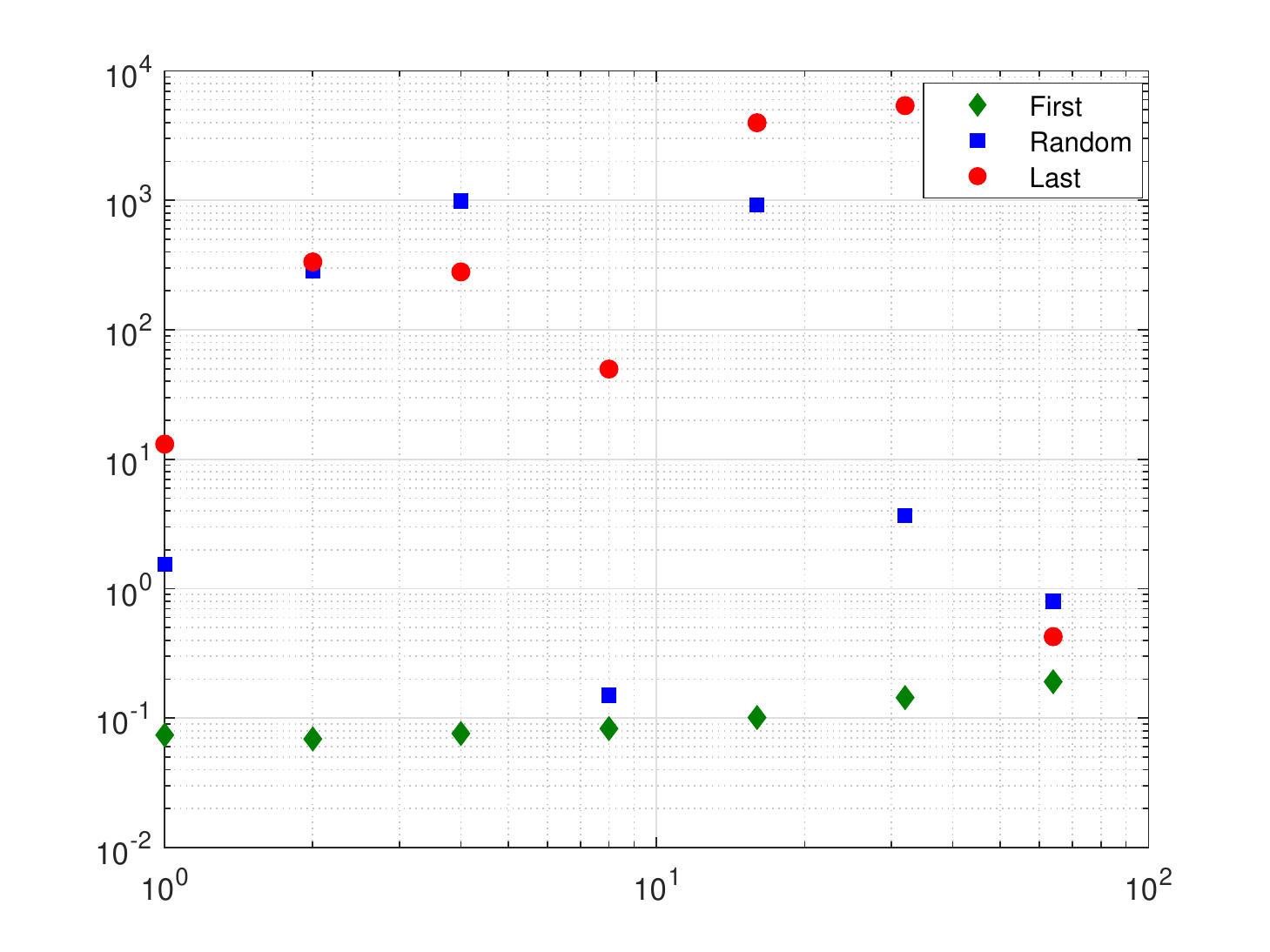}
    \end{minipage}
    \caption{Inference time (in s) for different size and depth of evidence for win95pts (left) and hepar2 (right) - log log scale}\label{fig3}
\end{figure}

\paragraph{The influence of observations.}
In order to answer the second question, we have chosen different ways to pick the set of evidences. This is aimed to investigate how the number of evidence nodes and their depth in the topological order affect the verification time. For each BN, three different sets of observations are considered; the evidence nodes at the beginning in the topological order, a random selection, and the last nodes in the topological ordering. We also varied the number of evidence nodes.
 Figure \ref{fig3} (in log-log scale) demonstrates the results for two large benchmarks, win95pts (left) and hepar2 (right). The x-axis denotes the number of evidence nodes and the y-axis denotes the model checking time in seconds. For the "first" setting, where $i$ nodes are picked from the beginning of the topological order, the time for performing model checking is relatively small; less than 3.064s seconds in all the experiments for win95pts and less than 0.682 seconds in all the experiments for hepar2. 
 The results follow a similar pattern in almost all the other BN benchmarks. The last nodes in the topological order are the highest dependent ones on the other nodes. That explains why model checking is significantly more time-consuming in the "last" setting. The verification time becomes negligible if the number of evidences is large. That is mostly because then the final result of the inference tends more likely to become zero when the number of evidence nodes are high. In this case there are many restrictions to be satisfied, and the zero probability can be computed very fast. 
 
\begin{table}
\newcolumntype{C}[1]{>{\centering\arraybackslash}m{#1}}
\centering
\caption{Empirical comparison with PSDD based inference regarding different vtree methods - Binary cases } \label{tab : 3}
    \begin{tabular}{|C{1.5cm}|C{0.75cm}|c|c|c|c|c|c|c|c|}
        \hline
        \multirow{8}{*}{\rotatebox{0}{BN}} & \multirow{9}{*}{\rotatebox{90}{\#Evidence }} & \multicolumn{2}{c|}{ MTBDD } & \multicolumn{2}{c|}{ \thead{PSDD \\ random vtree}  } & \multicolumn{2}{c|}{ \thead{PSDD \\ fixed vtree}  } & \multicolumn{2}{c|}{  \thead{PSDD \\ minfill vtree} } \\  \hline
        & & \rotatebox{90}{Compilation } & \rotatebox{90}{Inference }  & \rotatebox{90}{Compilation }  & \rotatebox{90}{Inference }  & \rotatebox{90}{Compilation }  & \rotatebox{90}{Inference }  & \rotatebox{90}{Compilation }  & \rotatebox{90}{Inference }  \\
        \hline \hline
        
        \multirow{5}{*}{ cancer } & 1 & \multirow{5}{*}{ 0.022s } & \textbf{0.001s} & \multirow{5}{*}{ 0.016s } & {0.002s} & \multirow{5}{*}{ 0.016s } & 0.003s & \multirow{5}{*}{ 0.005s } & {0.003s} \\
        & 2 & & \textbf{0.001s} & & 0.003s & & 0.003s & & {0.003s} \\
        & 3 & & \textbf{0.001s} & & 0.003s & & 0.003s & & {0.003s} \\
        & 4 & & \textbf{0.001s} & & 0.003s & & 0.002s & & 0.003s \\
        & 5 & & \textbf{0.001s} & & 0.003s & & 0.003s & & {0.003s} \\
        \hline
        
        \multirow{3}{*}{\thead{earth- \\quake}} & 1 & \multirow{3}{*}{ 0.006s } & \textbf{0.001s} & \multirow{3}{*}{ 0.015s } & {0.003s} & \multirow{3}{*}{ 0.016s } & {0.003s} & \multirow{3}{*}{ 0.004s } & {0.003s} \\
        & 2 & & \textbf{0.001s} & & {0.003s} & & {0.003s} & & {0.003s} \\
        & 4 & & \textbf{0.001s} & & {0.003s} & & {0.003s} & & {0.003s} \\
        \hline
        
        \multirow{4}{*}{ asia } & 1 & \multirow{4}{*}{ 0.018s } & \textbf{0.001s} & \multirow{4}{*}{ 0.026s } & 0.004s &
        \multirow{4}{*}{ 0.023s } & {0.003s} & \multirow{4}{*}{ 0.005s } & {0.003s} \\
        & 2 & & \textbf{0.001s} & & {0.003s} & & {0.003s}& & {0.003s} \\
        & 4 & & \textbf{0.001s} & & 0.004s & & 0.004s & & {0.003s} \\
        & 8 & & \textbf{0.002}s & & 0.003s & & {0.003s} & &{0.003s} \\
        \hline
        
        \multirow{7}{*}{ win95pts } & 1 & \multirow{7}{*}{ 11.214s } & 2.409s & \multirow{7}{*}{ 0.258s } & 0.074s & \multirow{7}{*}{ 0.233s } & 0.068s & \multirow{7}{*}{ 0.047s } & \textbf{0.042s} \\
        & 2 & & 2.760s & & 0.066s & & 0.060s & & \textbf{0.039s} \\
        & 4 & & 2.501s & & 0.067s & & 0.067s & & \textbf{0.039s} \\
        & 8 & & 2.452s & & 0.063s & & 0.061s & & \textbf{0.039s} \\
        & 16 & & 2.576s & & 0.056s & & 0.050s & & \textbf{0.033s} \\
        & 32 & & 0.671s & & {0.053s} & & 0.046s & & \textbf{0.033s} \\
        & 64 & & {0.658s} & & {0.053s} & & {0.043} & & \textbf{0.032s} \\
        \hline
        
        \multirow{8}{*}{ andes } & 1 & \multirow{8}{*}{ 180s } & \textbf{1.165s} & \multirow{8}{*}{ 12.617s } & 5.479s & \multirow{8}{*}{ 13.046s } & 12.893s & \multirow{8}{*}{ 4.724s } & 4.863s \\
        & 2 & & \textbf{0.989s} & & 5.824s & & 12.832s & & 4.818s \\
        & 4 & & \textbf{0.992s} & & 5.423s & & 13.312s & & 4.823s \\
        & 8 & & \textbf{1.144s} & & 5.620s & & 12.874s & & 4.838s \\
        & 16 & & \textbf{1.247s} & & 5.612s & & 9.921s & & 4.122s \\
        & 32 & & \textbf{1.385s} & & 5.457s & & 10.362s & & 4.120s \\
        & 64 & & \textbf{2.538s} & & 5.552s & & 8.996s & & 3.442s \\
        & 128 & & 3.488s & & 3.656s & & 8.096s & & \textbf{3.141s} \\
        \hline
    \end{tabular}
\end{table}

\begin{table}
\newcolumntype{C}[1]{>{\centering\arraybackslash}m{#1}}
    \centering
\caption{Empirical comparison with PSDD based inference regarding different vtree methods - Non-binary cases} \label{tab : 4}
    \begin{tabular}{|C{0.08\linewidth}|C{0.05\linewidth}|C{0.04\linewidth}|c|c|c|c|c|c|c|c|}
        \hline
        \multirow{8}{*}{\rotatebox{0}{BN} } & \multirow{10}{*}{\rotatebox{90}{\#Nodes} } & \multirow{9}{*}{\rotatebox{90}{\#Evidence }} & \multicolumn{2}{c|}{MTBDD} & \multicolumn{2}{c|}{\thead{PSDD\\ random vtree}} & \multicolumn{2}{c|}{\thead{PSDD \\ fixed vtree}} & \multicolumn{2}{c|}{\thead{PSDD \\ minfill vtree}} \\ 
        \hline
        & & & \rotatebox{90}{Compilation } & \rotatebox{90}{Inference } & \rotatebox{90}{Compilation } & \rotatebox{90}{Inference } & \rotatebox{90}{Compilation } & \rotatebox{90}{Inference } & \rotatebox{90}{Compilation } & \rotatebox{90}{Inference } \\
        
        \hline \hline
        \multirow{3}{*}{ survey } & \multirow{3}{*}{ 14 } & 1 & \multirow{3}{*}{ 0.018s } & \textbf{0.002s} & \multirow{3}{*}{ 0.114s } & 0.004s & \multirow{3}{*}{ 0.129s } & 0.004s & \multirow{3}{*}{ 0.019s } & 0.004s \\
        & & 2 & & \textbf{0.002s} & & 0.004s & & {0.003s} & & 0.004s \\
        & & 4 & & 0.004s & & \textbf{0.003s} & & \textbf{0.003s} & & \textbf{0.003s} \\
        \hline
        
        \multirow{4}{*}{ sachs } & \multirow{4}{*}{ 33 } & 1 & \multirow{4}{*}{ 0.076s } & \textbf{0.002s} & \multirow{4}{*}{ 0.208s } & {0.008s} & \multirow{4}{*}{ 0.212s } & 0.008s & \multirow{4}{*}{ 0.096s } & 0.010s \\
        & & 2 & & \textbf{0.002s} & & {0.008s} & & 0.008s & & {0.008s} \\
        & & 4 & & \textbf{0.004s} & & 0.009s & & 0.009s & & 0.009s \\
        & & 8 & & \textbf{0.004s} & & {0.008s} & & {0.007s} & & {0.008s}\\
        \hline

        \multirow{5}{*}{ child } & \multirow{5}{*}{ 60 } & 1 & \multirow{5}{*}{ 0.273s } & 0.014s & \multirow{5}{*}{ 0.304s } & 0.014s & \multirow{5}{*}{ 0.293s } & \textbf{0.012s} & \multirow{5}{*}{ 0.191s } & 0.021s \\
        & & 2 & & \textbf{0.004s} & & 0.013s & & {0.010s} & & 0.018s \\
        & & 4 & & \textbf{0.005s} & & 0.013s & & {0.010s} & & 0.018s \\
        & & 8 & & \textbf{0.006s} & & {0.010s} & & {0.010s} & & {0.013s} \\
        & & 16 & & \textbf{0.008s} & & 0.011s & & 0.010s & & 0.016s \\
        \hline

        \multirow{6}{*}{ alarm } & \multirow{6}{*}{ 104 } & 1 & \multirow{6}{*}{ 1.538s } & \textbf{0.010s} & \multirow{6}{*}{ 0.703s } & 0.014s & \multirow{6}{*}{ 0.685s } & 0.012s & \multirow{6}{*}{ 0.345s } & 0.013s \\
        & & 2 & & \textbf{0.011s} & & 0.012s & & 0.013s & & 0.012s \\
        & & 4 & & 0.013s & & 0.013s & & 0.013s & & \textbf{0.012s} \\
        & & 8 & & 0.014s & & \textbf{0.013s} & & \textbf{0.013s} & & \textbf{0.013s} \\
        & & 16 & & 0.019s & & \textbf{0.010s} & & 0.011s & & \textbf{0.010s} \\
        & & 32 & & 0.031s & & 0.013s & & 0.013s & & \textbf{0.012s} \\
        \hline

        \multirow{5}{*}{ \thead{insur- \\ ance} } & \multirow{5}{*}{ 88 } & 1 & \multirow{5}{*}{ 2.258s } & 0.432s & \multirow{5}{*}{ 0.695s } & \textbf{0.011s} & \multirow{5}{*}{ 0.672s } & 0.013s & \multirow{5}{*}{ 0.342s } & 0.012s \\
        & & 2 & & 0.462s & & \textbf{0.012s} & & \textbf{0.012s} & & \textbf{0.012s} \\
        & & 4 & & 0.461s & & \textbf{0.012s} & & 0.013s & & \textbf{0.012s}\\
        & & 8 & & 0.478s & & 0.013s & & 0.013s & & \textbf{0.012s} \\
        & & 16 & &{0.174s} & & 0.012s & & {0.011s} & & \textbf{0.010s} \\
        \hline
        
        \multirow{7}{*}{ \thead{he- \\ par2} } & \multirow{7}{*}{ 162 } & 1 & \multirow{7}{*}{ 17s } & 0.074s & \multirow{7}{*}{ 32.129s } & 0.058s & \multirow{7}{*}{ 37.466s } & \textbf{0.054s} & \multirow{7}{*}{ 12.205s } & 0.056s \\
        & & 2 & & {0.069s} & & {0.054s} & & \textbf{0.049s} & & {0.054s} \\
        & & 4 & & 0.076s & & 0.052s & & \textbf{0.044s} & & 0.053s \\
        & & 8 & & 0.083s & & 0.052s & & \textbf{0.043s} & & \textbf{0.043s} \\
        & & 16 & & 0.101s & & 0.045s & & 0.043s & & \textbf{0.033s} \\
        & & 32 & & 0.144s & & {0.039s} & & \textbf{0.036s} & & 0.051s \\
        & & 64 & & 0.191s & & 0.042s & & \textbf{0.038s} & & 0.051s \\
        \hline
    \end{tabular}
\end{table}

\paragraph{Comparing MTBDD-based model checking to inference using PSDDs.}
In order to answer our last question, we have conducted a series of experiments to see how our approach performs compared to the recent prominent inference tool based on PSDD. PSDD\footnote{\url{https://github.com/hahaXD/psdd}} is a scalable tool for reasoning and learning on belief networks in the AI literature. We have compiled the BNs in the benchmark into MTBDD with storm symbolic engine. To this end, we have converted the benchmark BNs into PSDDs and vtrees using the PSDD-Nips package\footnote{\url{https://github.com/hahaXD/psdd\_nips}}. Our experiments covers the available vtree methods: random, fixed and minfill. The decisive difference between these methods is the heuristics they employ to triangulate the DAG underlying a BN \cite{DBLP:conf/pgm/HopkinsD02}.
Due to the fact that the PSDD packages are inherently limited to perform inference only on BNs with binary variables, we categorize our results into two parts: \emph{binary} BNs and \emph{non-binary} BNs. Table \ref{tab : 3} indicates the results for the binary benchmarks. The results include the compilation time and inference time by different methods, taking the same sets of evidence nodes. In each row the minimum inference time is highlighted in bold face.
As inference in our tool is applicable to non-binary BNs, we have built a prototypical script to binarize non-binary networks such that they can be fed into the PSDD package. Table \ref{tab : 4} indicates the results for these non-binary benchmarks where $\#{Nodes}$ indicates the number of resulting binary variables. The pre-processing time for conversion and binarization is not included. Due to the large number of parameters in the benchmarks "hailfinder", "water", and "pathfinder" (see Table \ref{tab1}), these cases are computationally hard to binarize. Therefore, these cases are not included in Table \ref{tab : 4}.
\paragraph{} The main conclusions of our experimental results are:
\begin{enumerate}
    \item Inference using MTBDD-based symbolic model checking is competitive to BN-specific symbolic techniques like PSDD for small to large BNs.
    \item PSDD techniques outperform our MTBDD-based approach for very large and huge BNs.
    \item MTBDD-based inference is quite sensitive to the number and depth (in the topological order) of evidences.
\end{enumerate}

%%%%%%%%%%%%%%%%%%%%%%%%%%%%%%%%%%%%% Section 7 %%%%%%%%%%%%%%%%%%%%%%%%%%%%%%%%%%%%%%%%%%%%%
\section{Conclusions}
In this paper, we have investigated MTBDD-based symbolic probabilistic model checking to perform exact inference on Bayesian networks. We have translated Bayesian networks into Markov chains, and have reduced inference to computing reachability probabilities. Our prototypical tool chain built on top of storm \cite{DBLP:conf/cav/DehnertJK017} is evaluated on BNs from the \texttt{bnlearn} repository. We investigated several hypotheses to see which factors are affecting the inference time.
\\
Future work consists of optimizing our implementation and approach, and to consider other metrics on BNs, such as maximum a posteriori (MAP) and the most probable explanation (MPE). We also like to generalize our approach to recursive BNs \cite{DBLP:journals/amai/Jaeger01} or dynamic BNs \cite{DBLP:books/sp/Jensen01}, which bring respectively the notion of recursion and time on the table. Moreover, we believe that this work provides a good basis for performing probabilistic model checking on a broader set of graphical models such as Markov networks, which are, unlike Bayesian networks, undirected in nature. 

\paragraph{Acknowledgement. }The authors would like to thank Yujia Shen (UCLA) for his kind support with running the PSDD tools.


\bibliographystyle{splncs04}
\bibliography{main}

\end{document}